\documentclass[10pt, conference, letterpaper]{IEEEtran}
\usepackage{geometry}
\AtBeginDocument{\geometry{top=0.75in, bottom=0.80in, left=0.75in, right=0.75in}}
\IEEEoverridecommandlockouts    

\usepackage{makecell}

\usepackage{amsmath,amsfonts}
\usepackage{dsfont}
\usepackage{algorithm}
\usepackage{algpseudocode}
\usepackage{hyperref}
\usepackage{cleveref}
\usepackage{array}
\usepackage{courier}

\hypersetup{hidelinks}
\usepackage{threeparttable} 
\usepackage{subfigure}
\usepackage{textcomp,soul}
\usepackage{stfloats}
\usepackage{url}
\usepackage{verbatim}
\usepackage{graphicx}
\usepackage{booktabs}
\usepackage{multirow}
\usepackage{eqparbox}
\usepackage{cite}
\usepackage{amsthm}
\usepackage{bm}
\usepackage{amssymb}
\usepackage{xspace}
\usepackage[export]{adjustbox}

\newcommand{\eg}{\textit{e.g.}\xspace}
\newcommand{\ie}{\textit{i.e.}\xspace}

\newcolumntype{C}[1]{>{\centering\arraybackslash}p{#1}}
\theoremstyle{definition}

\usepackage[table]{xcolor}
\definecolor{newpurple}{RGB}{165,154,202}
\definecolor{lightgreen}{rgb}{0.7,1,0.7}
\definecolor{newpink}{RGB}{244, 179, 194}
\definecolor{lightyellow}{RGB}{253, 244, 201}
\definecolor{newlightgreen}{RGB}{198, 224, 179}
\definecolor{mediumgray}{rgb}{0.55, 0.55, 0.55}
\definecolor{lightgray}{rgb}{0.75, 0.75, 0.75}
\definecolor{lightsalmon}{RGB}{248, 214, 190}
\definecolor{lightnewpurple}{RGB}{200, 176, 200}

\makeatletter
\pretocmd\@bibitem{\color{black}\csname keycolor#1\endcsname}{}{\fail}
\newcommand\citecolor[1]{\@namedef{keycolor#1}{\color{blue}}}
\makeatother

\title{
  \vspace{1em} \LARGE \bf Decision-Making with Lightweight Confidence-Aware Language Model for Autonomous Driving}

\author{
	\parbox{\textwidth}{%
		\centering
		Ruoyu Yao, Ruiguo Zhong,  Pei Liu, Mingxing Peng, Rui Yang, Jun Ma$^{\dagger}$%
	}%
	\thanks{Ruoyu Yao, Ruiguo Zhong, Pei Liu, Mingxing Peng, Rui Yang, and Jun Ma are with The Hong Kong University of Science and Technology (Guangzhou), Guangzhou 511453, China (e-mail: \{ryao092, rzhong151, pliu061, mpeng060, ryang253\}@connect.hkust-gz.edu.cn; jun.ma@ust.hk).}%
	\thanks{$^{\dagger}$Corresponding author.
}
}

\begin{document}
\maketitle
\thispagestyle{empty}
\pagestyle{empty}

\begin{abstract}
Large Language Models (LLMs) and Multimodal LLMs (MLLMs) have demonstrated immense potential in autonomous driving (AD) by offering human-like reasoning and open-world generalization. However, the excessive computational overhead and high inference latency of these massive models severely hinder their deployment in resource-constrained AD systems. To address this challenge, we propose a novel decision-making framework utilizing a lightweight confidence-aware language model, which bridges the gap between complex multimodal intention reasoning and efficient inference. Specifically, we design a multi-agent collaborative workflow, comprising action voting, confidence assessment, and summarization agents, to generate high-quality, confidence-annotated decision demonstrations via explicit Chain-of-Thought (CoT) reasoning. These demonstrations are then distilled into a lightweight language model featuring a dual-head architecture, enabling the joint prediction of decision probabilities and the generation of textual rationales. The distillation is realized via a confidence-aware fine-tuning strategy coupled with Retrieval Augmented Generation (RAG) to enhance the model's adaptability and data efficiency. Comprehensive closed-loop experiments on the nuPlan benchmark demonstrate that our approach achieves state-of-the-art (SOTA) success rates in both regular and long-tail scenarios while maintaining low inference latency.
\end{abstract}

\section{Introduction} \label{sec:intro}
Decision-making acts as the cognitive “brain” of autonomous driving (AD) systems, interpreting the perceived environment to generate safe, rational, and compliant high-level driving behaviors that guide motion planning and control~\cite{Brian2016asurvey, Zhang2024Active}. Current research on decision-making prioritizes learning-based methods  to enhance robustness and generalizability beyond handcrafted rules~\cite{li2021safe, yang2025interactive}. Recently, AD systems are increasingly leveraging Large Language Models (LLMs) or Multimodal LLMs (MLLMs) to achieve human-level decision-making capabilities~\cite{wendilu, mei2024contin, fang2025towards}. These models offer unprecedented potential for handling complex, open-world scenarios through their vast knowledge bases and reasoning abilities~\cite{cui2024survey}.

Despite the advancements, most large foundation model empowered AD systems are limited to deterministic patterns. They typically use LLMs to infer decisions or trajectories via in-context learning~\cite{wendilu, sha2023languagempc} or supervised learning~\cite{shao2024lmdrive, jiang2024senna, liu2025dsdrive}, often neglecting explicit uncertainty representation~\cite{cai2020vtgnet}. Consequently, the learned policy either settles on a locally optimal behavior or collapses into an infeasible interpolation between multiple valid intentions~\cite{lee2017desire}. This undermines the very promise of large foundation models to encode rich world knowledge and support robust, multimodal reasoning in open-world decision-making.

To capture the stochastic nature of driving intentions, recent efforts integrate multimodal driving policy into the auto-regressive generation framework. In~\cite{yao2024calmm}, a confidence-aware decision-making and motion planning framework is proposed. The decision-making module built upon an MLLM is employed for Top-K confident driving decision reasoning, followed by a diffusion-based multimodal motion planning with a scoring and selection mechanism to determine the best driving strategy. In~\cite{fu2025orion}, an end-to-end AD architecture is presented, which adopts a variational autoencoder (VAE) to bridge the MLLM's reasoning and action space for generating multimodal planning trajectories. These frameworks demonstrate the capability of generating both flexible and realistic multimodal driving maneuvers, attaining superior closed-loop performance over the deterministic counterpart. However, the excessive computational costs incurred by the auto-regressive reasoning process significantly challenge their real-world applicability. This motivates a lightweight approach that preserves diverse intention reasoning and uncertainty quantification in multimodal decision-making, while enabling substantially lower inference complexity. 

We propose a novel decision-making framework for AD based on a lightweight language model that bridges the gap between confidence-aware multimodal intention reasoning and efficient inference. The core insight lies in a confidence-aware distillation scheme grounded in the paradigm of slow-fast systems~\cite{kahneman2011fast, mei2024contin}. In this scheme, multiple collaborative agents instantiated by LLMs are employed to generate high-quality, confidence-aware decision-making demonstrations through slow and explicit Chain-of-Thought (CoT) reasoning. Then, a lightweight language model is leveraged to learn from the demonstrations and gain the capability for fast inference \textit{before} completing language explanations. Only the lightweight model is required for on-board deployment, reducing both memory usage and inference latency. Our contributions are summarized as follows:
\begin{itemize}
    \item We propose a multimodal decision-making framework for AD based on a lightweight language model. It features a dual-head architecture, enabling joint prediction of decision probabilities and generation of textual rationales, thereby supporting efficient and interpretable multimodal decision-making.
    \item We develop a multi-agent collaborative reasoning workflow composed of specialized LLM agents, designed to obtain confidence-aware decision-making demonstrations. Each agent is assigned a role-specific CoT reasoning subtask, enabling the system to generate plausible driving strategies with confidence estimates.
    \item We present a confidence-aware fine-tuning strategy combined with Retrieval Augmented Generation (RAG). This synthesis of supervised and in-context learning enables the lightweight model to learn from demonstrations with high data efficiency.
    \item Comprehensive experiments on the nuPlan benchmark  demonstrate that our approach achieves state-of-the-art (SOTA) success rates in both regular and long-tail scenarios, attaining closed-loop scores comparable to computationally intensive large model-based methods.
\end{itemize}

\begin{figure*}
  \centering
    \includegraphics[width=0.85\linewidth]{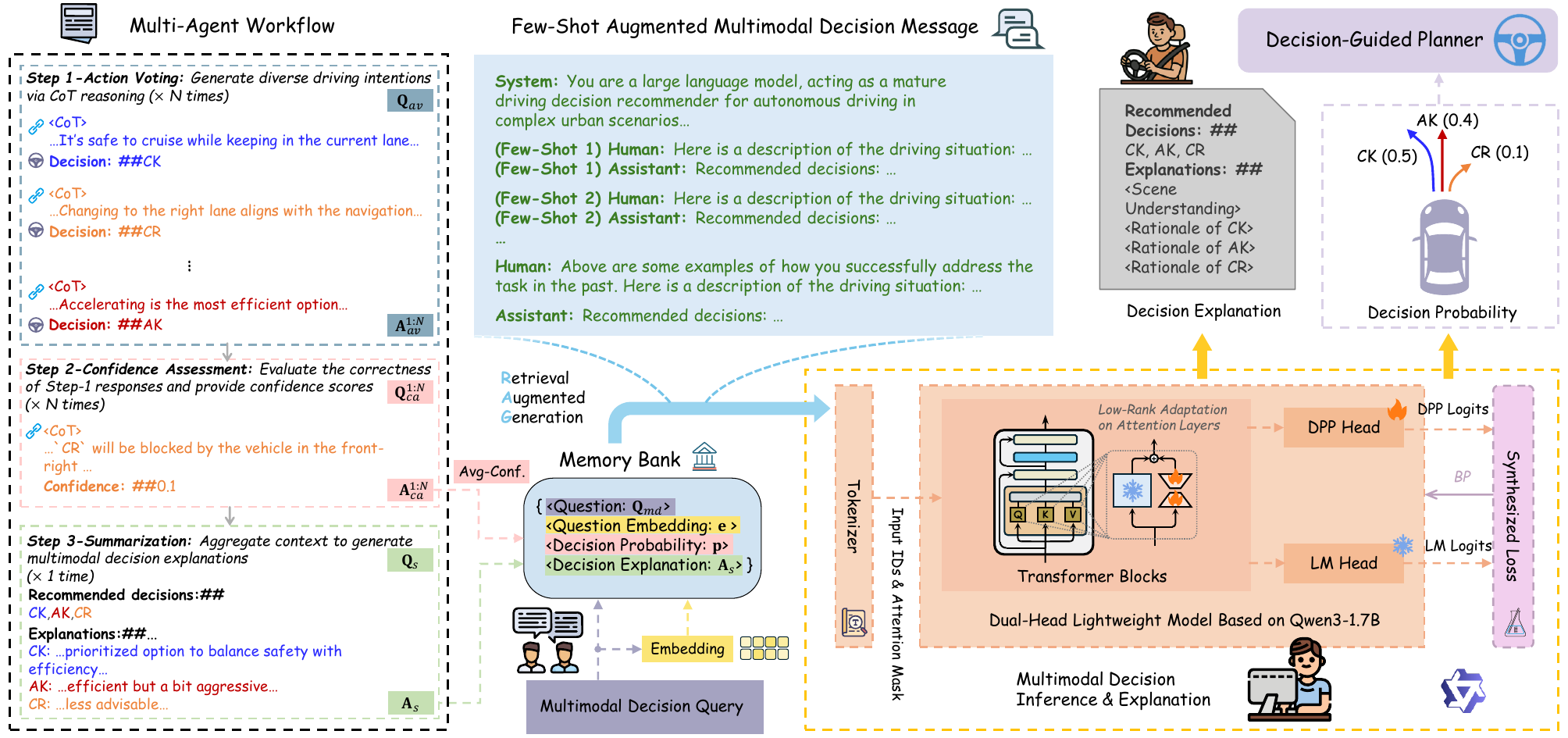}
  \caption{An illustration of our framework, which mainly consists of a multi-agent workflow for collecting memories featuring confidence-aware multimodal decision-making, and a dual-head lightweight language model synthesizing decision probability prediction (DPP) and language modeling (LM) via supervised fine-tuning combined with RAG. Decision probabilities are sent to a downstream planning module, guiding multimodal trajectory planning and scoring.}
  \label{fig:model_framework}
\vspace{-0.4cm}
\end{figure*}


\section{Related Works}

\subsection{Knowledge-Driven Autonomous Driving}
The integration of LLMs has shifted autonomous driving from data-driven imitation to knowledge-driven reasoning~\cite{wendilu, cui2024survey}. Pioneering works like GPT-Driver~\cite{Mao2023GPTDriver} utilize LLMs to comprehend traffic semantics and generate driving actions with textual rationales. To enable continuous improvement, subsequent works~\cite{wendilu, Fu2024DriveLikeHuman} integrate memory modules for adapting to open-world scenarios via few-shot learning. However, the substantial inference latency of these parameter-heavy models remains a critical bottleneck for deployment on vehicle hardware~\cite{Xu2024Survey}. This limitation has motivated the investigation of dual-process architectures inspired by human cognitive processes~\cite{kahneman2011fast, mei2024contin, ma2025leapvad}, where a slow, deliberative teacher model guides a fast, reactive policy.

\subsection{Multimodal Decision-Making and Motion Planning}
Modeling uncertainty is essential to prevent ``mode collapse'' in complex scenarios~\cite{lee2017desire}. Generative planning methods based on VAE~\cite{zheng2024genad} and diffusion models~\cite{zhengdiffusion, zhong2025coplanner}, have achieved success in modeling multimodal future distributions. However, vanilla generative planners lack the ability to perform high-level, socially compliant reasoning over traffic scenarios, thereby producing unrealistic driving behaviors in unseen situations. Although recent studies integrate large foundation models with generative planners to bridge common-sense reasoning and multimodal motion planning~\cite{wang2024he, yao2024calmm, fu2025orion}, their reliance on the auto-regressive inference of large models incurs excessive computational costs, limiting high-frequency applicability.

\subsection{Knowledge Distillation for Efficient Inference}
Knowledge distillation offers a viable path to reconcile reasoning depth with inference speed~\cite{Xu2024Survey, hinton2015distilling}. 
Foundational works emphasized the use of soft labels to transfer the teacher's underlying uncertainty and ``dark knowledge'', preventing the information loss inherent in one-hot targets~\cite{gou2021knowledge}. 
In the era of LLMs, this concept has expanded to the distillation of reasoning capabilities using content generated by the teacher model, such as CoT rationales~\cite{hsieh2023distilling}. In autonomous driving, recent works employ teacher-student paradigms to capture multimodal driving strategies~\cite{li2024hydra, yu2025distilldrive} or condense complex reasoning into reactive policies~\cite{mei2024contin, liu2025dsdrive}. Distinct from existing studies limited to a single objective, our dual-head lightweight model learns the decision probability distribution alongside textual rationales, inheriting both the uncertainty of classic distillation~\cite{hinton2015distilling, zhou2024enhancing} and the interpretability of large foundation models~\cite{hsieh2023distilling}.


\section{Methodology}\label{sec:method}
The overall framework is shown in Fig.~\ref{fig:model_framework}. It employs a multi-agent workflow for generating confidence-aware decision-making demonstrations, a lightweight language model that distills from the demonstrations and ensures efficient inference, and a decision-guided motion planner to produce planning trajectories.

\subsection{Multi-Agent Workflow for Confidence-Aware Decision-Making Demonstrations}
To obtain diverse intention reasoning paths and probabilities of their correctness, we tailor a multi-agent workflow that takes task-specific descriptions as input and conducts a three-stage collaborative reasoning process: \textit{action voting}, \textit{confidence assessment}, and \textit{summarization}. This decomposes the complicated multimodal driving decision-making problem into a series of tractable and well-defined subtasks, each of which can be effectively handled by a specialized agent. It also liberates confidence elicitation from a fixed line of thinking~\cite{xiongcan}. Operating solely during data collection, the workflow preserves on-board inference efficiency.

\subsubsection{Action Voting Agent}
To simulate diverse human-like decision-making process, we instantiate $N$ independent action voting agents $\{\mathcal{V}^{(i)}\}_{i=1}^N$ by repeatedly prompting the same LLM with identical \textbf{driving context} descriptions and \textbf{prior knowledge}, while adopting stochastic decoding with a fixed temperature $\tau > 0$ and Top-K sampling. Each agent produces its own reasoning trace and action choice due to the inherent randomness in the sampling process.

Following~\cite{yao2024calmm}, we construct scenario-based prompts incorporating road structures, ego vehicle, surrounding vehicles and vulnerable road users (VRUs), static objects, traffic light states, and navigation commands to convey essential driving context. Then we inject essential prior knowledge into action voting agents, encompassing the definition of coordinate system, the definition of action space, and the reasoning guideline. An ego-centric polar coordinate system is leveraged to clearly represent the distance and angular relationships between objects. The action space $\mathcal{A}$ is the Cartesian product of longitudinal (acceleration, deceleration, cruise, stop) and lateral (left/right lane-change, lane keep) actions. Finally, the reasoning guideline requests each agent to perform CoT reasoning following three explicit steps to derive the driving action: \textit{holistic scene understanding}, \textit{key object identification}, and \textit{action selection}. The instruction message for each step is detailed in \Cref{tab:actvot_guideline}. 

Canonically, the action voting stage is formalized as:
\begin{subequations}
\begin{align}
\mathbf{Q}_{av} &= \mathcal{P}^{av}\text{-map}(\mathbf{s}_t, \mathbf{K}_{av}),\\
\mathbf{R}_{1}^{(i)} &= \mathcal{G}_1\big(\mathbf{Q}_{av};\ \tau\big), \label{eq:vot_r1} \\
\mathbf{R}_{2}^{(i)} &= \mathcal{G}_2\big(\mathbf{Q}_{av}, \mathbf{R}_{1}^{(i)};\ \tau\big), \label{eq:vot_r2} \\
\mathbf{R}_{3}^{(i)} &= \mathcal{G}_3\big(\mathbf{Q}_{av}, \mathbf{R}_{1}^{(i)}, \mathbf{R}_{2}^{(i)};\ \tau\big), \label{eq:vot_r3} \\
\mathbf{a}_t^{(i)} &= \mathcal{A}\text{-map}\big(\mathbf{R}_{3}^{(i)}\big) \in \mathcal{A}, \label{eq:vot_action}\\
\mathbf{A}_{av}^{(i)} &= \mathbf{R}_{1}^{(i)} \,\Vert\,  \mathbf{R}_{2}^{(i)} \,\Vert\,  \mathbf{R}_{3}^{(i)}
\end{align}
\label{eq:action_voting}
\end{subequations}
where \(\mathbf{s}_t\) denotes the driving context at time-step \(t\), and \(\mathbf{K}_{av}\) denotes the prior knowledge required by the action voting agent. \(\mathcal{P}^{av}\text{-map}\) stands for the deterministic mapping that transfers the original information into a prompt question \(\mathbf{Q}_{av}\). \(i \in \{1,...,N\}\) indexes the \(i\)-th voting agent. \(\mathcal{G}_1\), \(\mathcal{G}_2\), and \(\mathcal{G}_3\) denote the generation process of LLM corresponding to the three reasoning step, with \(\mathbf{R}_{1}^{(i)}\), \(\mathbf{R}_{2}^{(i)}\), and \(\mathbf{R}_{3}^{(i)}\) representing the generated contents. \(\mathcal{A}\text{-map}\) is a deterministic parser that extracts the final discrete action from \(\mathbf{R}_{3}^{(i)}\). The generated contents of each step constitute the full answer \(\mathbf{A}_{av}^{(i)}\).

\begin{table*}[htbp]
\centering
\resizebox{0.9\textwidth}{!}{%
\begin{minipage}{\textwidth}
\caption{Reasoning Guideline for Action Voting.}
\label{tab:actvot_guideline}
\renewcommand{\arraystretch}{1.0} 
\begin{tabular}{>{}p{3.0cm}p{\dimexpr\textwidth-3.0cm-4\tabcolsep}}
\toprule
\multicolumn{1}{l}{\textbf{Reasoning Step}} & \multicolumn{1}{l}{\textbf{Instruction Message}} \\
\midrule
Step 1 - Holistic Scene Understanding & 
\textit{Analyze the driving context, including the road structure, traffic light, navigation for the ego vehicle (if applicable), and the movement of surrounding vehicles, describe their implications on the ego vehicle's behavior.} \\
\midrule
Step 2 - Key Object Identification & 
\textit{Identify the most important objects affecting the ego vehicle's behavior (vehicles, VRUs, or static objects). Focus only on critical ones, such as:
             vehicles with lane conflicts, \eg, vehicles in the same lane, vehicles intending to switch into the ego lane, vehicles in the lane the ego vehicle will switch to; vehicles at junctions and close ahead of the ego vehicle; VRUs within 15 m, with a speed and heading angle suggesting potential collision;Static objects directly obstructing the ego vehicle's path.} \\
\midrule
Step 3 - Action Selection & 
\textit{Based on the step 1 \& 2 analysis and common-sense knowledge, predict the best decision from the \{action\_set\}.} \\
\bottomrule
\end{tabular}
\end{minipage}%
}
\vspace{-0.4cm}
\end{table*}

\subsubsection{Confidence Assessment Agent}
For each of the $N$ action voting agents, a dedicated confidence assessment instance is invoked to evaluate the correctness of its individual reasoning path. To organize the confidence assessment question \(\mathbf{Q}_{ca}\), we leverage the question \(\mathbf{Q}_{av}\) and answer \(\mathbf{A}_{av}^{(i)}\) of the action voting agent as the context and impose an additional instruction: \textit{How likely is the above decision-making process to be correct? Analyze the response, provide your reasoning concisely, and give your confidence (between 0.0 and 1.0) in this response}. Formally, the confidence for the $i$-th voting agent is obtained via:
\begin{subequations}
    \begin{align}
        \mathbf{Q}^{(i)}_{ca} &= \mathcal{P}^{ca}\text{-map}(\mathbf{Q}_{av}, \mathbf{A}_{av}^{(i)}),\\
        \mathbf{A}_{ca}^{(i)} &= \mathcal{G}_4\big(\mathbf{Q}^{(i)}_{ca};\tau_0\big), \\
         c^{(i)} &= \mathcal{C}\text{-map} \big(\mathbf{A}_{ca}^{(i)}\big)  \in \left[0.0, 1.0\right],
    \end{align}
\end{subequations}
where $\mathcal{P}^{ca}\text{-map}$ denotes the organization of the confidence assessment question. \(\mathcal{G}_4\) represents the generation of the response for confidence assessment, with \(\tau_0\) denoting a constant temperature of zero. The confidence of the \(i\)-th action voting response, \(c^{(i)}\), is parsed from the answer content \(\mathbf{A}_{ca}^{(i)}\). This per-agent assessment not only reduces context-length pressure but also mitigates anchoring bias that may arise when an LLM observes a majority vote before assigning confidence~\cite{zhu2025conformity}.

\subsubsection{Summarization Agent and Confidence Aggregation}
In the last stage, we leverage a summarization agent and an aggregation operation to obtain both the textual explanations and probabilities of chosen actions. Taking the full context of prior question-answering records as input, the summarization agent aggregates diverse reasoning paths to generate recommended decisions with rationales. It is requested not to leak the existence of other agents or the processes in the workflow, such that the lightweight language model is enforced to capture multimodal decision-making patterns in an individual reasoning manner. The summarization process is formulated as:
\begin{subequations}
\begin{align}
    \mathbf{Q}_{s} &= \mathcal{P}^{s}\text{-map}(\mathbf{Q}^{(1:N)}_{ca}, \mathbf{A}_{ca}^{(1:N)}),\\
    \mathbf{A}_{s} &= \mathcal{G}_5\big(\mathbf{Q}_{s};\tau_0\big),\label{eq:sum}
\end{align}
\end{subequations}
where \(\mathbf{Q}_{s}\) and \(\mathbf{A}_{s}\) represent the question and answer in this stage, respectively. Then, we average out the confidence scores of different discrete actions to obtain normalized decision probabilities:
\begin{equation}
    p(a \mid \mathbf{s}_t, \mathbf{K}_{md}) = 
    \frac{1}{Z} \sum_{i=1}^{N} c^{(i)} \cdot \mathbb{I}\big( \mathbf{a}_t^{(i)} = a \big),
    \quad \forall a \in \mathcal{A},
    \label{eq:conf_agg}
\end{equation}
where $\mathbb{I}(\cdot)$ is the indicator function, and $Z = \sum_{a' \in \mathcal{A}} \sum_{i=1}^{N} c^{(i)} \cdot \mathbb{I}\big( \mathbf{a}_t^{(i)} = a' \big) = \sum_{i=1}^N c^{(i)}$ is the normalization constant ensuring $\sum_{a \in \mathcal{A}} p(a \mid \mathbf{s}_t, \mathbf{K}_{md}) = 1$. \(\mathbf{K}_{md}\) denotes the prior knowledge required for multimodal decision-making. The only difference between \(\mathbf{K}_{av}\) and \(\mathbf{K}_{md}\) is that the latter does not incorporate an explicit reasoning guideline, as the capability of multimodal intention reasoning is internalized via supervised fine-tuning and reinforced by RAG.

\subsubsection{Memory Accumulation}
To finish the collection of a memory item, we construct the multimodal decision-making question by integrating the driving context and prior knowledge: \(\mathbf{Q}_{md}=\mathcal{P}^{md}\text{-map}(\mathbf{s}_t, \mathbf{K}_{md})\). This is followed by a text embedding operation to map the driving context description to a high-dimensional vector \(\mathbf{e}\),  which allows the similarity assessment in the retrieval process. Incorporating the textual explanation from Eq.~\eqref{eq:sum} and decision probabilities from Eq.~\eqref{eq:conf_agg}, each memory item is represented as a tuple of \(<\mathbf{Q}_{md}, \mathbf{e}, \mathbf{p}, \mathbf{A}_s>\).

\subsection{Lightweight Language Model for Efficient Inference}
To enable low-latency inference, we distill the confidence-aware decision-making patterns from the multi-agent workflow into a lightweight language model.

\subsubsection{RAG Mechanism}
During the training process, we augment the input to the lightweight language model via RAG with a randomly sized set of few-shot examples. 
Specifically, for each training instance, we sample the number of retrieved examples $K$ uniformly from a predefined integer interval and  retrieve the Top-K most similar entries from the memory bank $\mathcal{D}$ based on embedding similarity:

\begin{small}
    \begin{subequations}
    \begin{align}
        K &\sim \mathcal{U}\big(\{K_{\min},\, K_{\min}+1,\, \dots,\, K_{\max}\}\big),
        \label{eq:rag_K} \\
        \mathcal{M}_K &= \{j_1, j_2, \dots, j_K\},
        \label{eq:rag_MK} \\
        \text{where} \quad
        j_m &= \arg\max_{j \in \mathcal{D} \setminus \{j_1,\dots,j_{m-1}\}} 
        \frac{\mathbf{e}^\top \mathbf{e}^{(j)}}{\|\mathbf{e}\|\,\|\mathbf{e}^{(j)}\|},
        \;\; m \in \{1,\dots,K\}.
        \nonumber
    \end{align}
    \label{eq:rag_process}
    \end{subequations}
\end{small}
$K_{\min} \ge 0$ and $K_{\max} > 0$ are hyperparameters, and $\mathcal{U}(\cdot)$ denotes the discrete uniform distribution over a finite set. Then the augmented query is constructed by concatenating these examples with the current query.
 \begin{equation}
         \tilde{\mathbf{Q}}_{md} 
    = \big\Vert_{m=1}^{K}
      \Big[
        \mathbf{Q}_{md}^{(j_m)} \,\Vert\, \mathbf{A}_s^{(j_m)}
      \Big]
      \;\Vert\;
      \mathbf{Q}_{md}.
 \end{equation}
Randomizing \(K\) during training exposes the model to diverse context lengths, mimicking real-world inference constraints imposed by memory or latency limits.

\subsubsection{Confidence-Aware Dual-Task Fine-Tuning}
\label{subsec:dual_task}
To equip the lightweight model with dual capabilities of decision probability prediction and textual rationale generation, we propose a dual-task fine-tuning framework. As illustrated in Fig.~\ref{fig:model_framework}, the model takes $\tilde{\mathbf{Q}}_{md}\,\Vert\,\mathbf{A}_s$ as input and learns two complementary objectives through parameter-efficient adaptation. 

Given the tokenized input sequence $\mathbf{x} = [x_1, x_2, \dots, x_M]$, the base language model $f_{\theta}$ processes the sequence to generate hidden states $\mathbf{H} = [\mathbf{h}_1, \mathbf{h}_2, \dots, \mathbf{h}_M]$, where $M$ denotes the sequence length. Crucially, we identify the \textit{prompt termination position} $m^*$ where decision probability prediction should occur, immediately before the assistant's response begins, \ie,
\begin{equation}
\begin{split}
m^* &= \max \left\{ m\;\middle|\; x_m = \text{\texttt{<|im\_start|>}}\right. \\
&\quad \left. \land x_{m+1} = \text{\texttt{assistant}} \right\} - 1.
\end{split}
\end{equation}

The dual-task learning proceeds as follows:

\textbf{Language Modeling Task.} The model performs next-token prediction over the input sequence. The loss is computed across all tokens in the sequence ($m \in \mathcal{T}$): 
\begin{equation}
    \mathcal{L}_{\text{LM}} = -\frac{1}{|\mathcal{T}|}\sum_{m \in \mathcal{T}} \log p(x_{m+1} \mid \mathbf{x}_{1:m};\theta).
\end{equation}

\textbf{Decision Probability Prediction Task.} The hidden state $\mathbf{h}_{m^*}$ at position $m^*$ is fed through a lightweight classifier head $g_{\phi}$ to predict decision probabilities:
\begin{equation}
    \hat{\mathbf{p}} = \text{softmax}\big(g_{\phi}(\mathbf{h}_{m^*})\big) \in \mathbb{R}^{10},
\end{equation}
where the classifier is trained to match the aggregated confidence distribution $p(a \mid \mathbf{s}_t, \mathbf{K}_{md})$ from Eq.~\eqref{eq:conf_agg}. Using KL divergence preserves the probabilistic calibration of confidence estimates:
\begin{equation}\label{eq:kld}
    \mathcal{L}_{\text{DPP}} = \text{KL}\Big( p(a \mid \mathbf{s}_t, \mathbf{K}_{md}) \parallel \hat{\mathbf{p}} \Big).
\end{equation}
The total training objective combines both tasks with a balancing factor $\lambda$:
\begin{equation}
\mathcal{L}_{\text{total}} = \mathcal{L}_{\text{LM}} + \lambda \mathcal{L}_{\text{DPP}}.
\end{equation}
During inference, the model outputs a distribution $\hat{\mathbf{p}}$ alongside textual rationale $\hat{\mathbf{A}}_s$. The former provides a decision-level probabilistic signal to inform downstream trajectory planning.

\subsection{Decision-Guided Motion Planning}
Similarly to \cite{yao2024calmm}, we utilize a diffusion-based motion planner \cite{yang2024diffusion} for decision-guided trajectory generation, followed by a confidence-aware scorer to arbitrate the best trajectory balancing high-level decision confidence and low-level trajectory quality. However, rather than fixing the number of candidate actions, we dynamically construct a candidate action set by retaining only those actions whose predicted probabilities exceed a confidence threshold:
\begin{equation}
\mathcal{C} = \big\{ a \in \mathcal{A} \; \big| \; \hat{p}(a\mid \mathbf{s}_t, \mathbf{K}_{md}) \ge \gamma_c \big\},
\end{equation}
where $\gamma_c \in (0,1)$ is a tunable confidence threshold. The size of $\mathcal{C}$ is thus adaptive to situational uncertainty, \ie, compact in deterministic scenarios and expanded in ambiguous ones.

\begin{table*}[ht]
    \centering
    \resizebox{0.85\textwidth}{!}{%
    \begin{minipage}{\textwidth}
    \centering
    \caption{Quantitative closed-loop performance of different approaches. All metrics are \underline{higher the better}. Best: \colorbox{newlightgreen}{green}, second best: \colorbox{lightyellow}{yellow}. \dag: Methods with rule-based modules for emergency braking. \(*\): Methods with results quoted from original publications. CALMM-Drive-L and CALMM-Drive-M are respectively powered by GPT-4o and GPT-4o-Mini.}
    \vspace{-0.1cm}
    \footnotesize 
    \begin{tabular}{@{}p{2.0cm}p{2.82cm}cccccccc@{}}
        \toprule
        \multirow{2}{*}{\textbf{Type}} & \multirow{2}{*}{\textbf{Planner}} & \multicolumn{4}{c}{\textbf{Test14-Hard}} & \multicolumn{4}{c}{\textbf{Test14-Random}} \\
        \cmidrule(lr){3-6} \cmidrule(lr){7-10}
        & & \textbf{NR-CLS}  & \textbf{NR-SR(\%)}  & \textbf{R-CLS}  & \textbf{R-SR(\%)}  & \textbf{NR-CLS}  & \textbf{NR-SR(\%)}  & \textbf{R-CLS}  & \textbf{R-SR(\%)} \\ 
        \midrule
        \textcolor{mediumgray}{Expert} & \textcolor{mediumgray}{Log-Replay} & \textcolor{mediumgray}{85.96} & \textcolor{mediumgray}{91.18} & \textcolor{mediumgray}{68.80} & \textcolor{mediumgray}{73.53} & \textcolor{mediumgray}{94.03} & \textcolor{mediumgray}{96.93} & \textcolor{mediumgray}{75.86} & \textcolor{mediumgray}{78.16}\\
        \midrule
        \multirow{2}{*}{Rule-Based} & IDM~\cite{helbing1998generalized} & 56.16 & 65.81 & 62.26 & 71.69 & 70.39 & 77.01 & 72.42 & 78.54 \\ 
        & PDM-Closed\(^{\dag}\)~\cite{dauner2023parting} & 65.08 & 78.68 & 75.19 & 85.66 & 90.05 & 94.64 & \makecell{\colorbox{newlightgreen}{91.64}} & \makecell{\colorbox{newlightgreen}{96.55}} \\ 
        \midrule
        \multirow{4}{*}{Data-Driven}             & UrbanDriverOL~\cite{scheel2022urban} & 51.67 & 59.20 & 49.06 & 55.51 & 63.57 & 70.88 & 60.92 & 66.28 \\
                                                 & RasterModel~\cite{caesar2021nuplan} & 50.65 & 58.09 & 52.44 & 59.56 & 67.70 & 73.56 & 68.64 & 74.33 \\
                                                 & PlanTF~\cite{cheng2024rethinking} & 72.56 & 79.04 & 60.34 & 67.28 & 85.60 & 90.80 & 78.86 & 85.44 \\
                                                 & Diffusion-Planner\cite{zhengdiffusion} & 75.29 & 82.72 & 68.83 & 78.31 & 88.98 & 93.87 & 83.21 & 90.42 \\
        \midrule
        \multirow{4}{*}{Hybrid} & PDM-Hybrid\(^{\dag}\)~\cite{dauner2023parting} & 65.99 & 79.41 & 76.07 & 86.40 & \makecell{\colorbox{lightyellow}{90.10}} & 95.02 & \makecell{\colorbox{lightyellow}{91.29}} & \makecell{\colorbox{lightyellow}{96.17}} \\
                                         & GameFormer~\cite{huang2023gameformer} & 62.94 & 74.26 & 61.53 & 72.79 & 76.89 & 85.29 & 77.80 & 86.59 \\
                                         & Diffusion-ES~\cite{yang2024diffusion} & \makecell{\colorbox{lightyellow}{77.54}} & 84.93 & \makecell{\colorbox{lightyellow}{77.75}} & \makecell{\colorbox{lightyellow}{87.13}} & 87.70 & 94.25 & 87.18 & 93.87 \\
                                         & PLUTO\(^{\dag}\)~\cite{cheng2024pluto} & \makecell{\colorbox{newlightgreen}{79.19}} & 88.24 & 75.75 & 84.19 & \makecell{\colorbox{newlightgreen}{91.93}} & \makecell{\colorbox{lightyellow}{95.79}} & 89.94 & \makecell{\colorbox{lightyellow}{96.17}} \\
        \midrule
        \multirow{4}{*}{Knowledge-Driven} & PlanAgent\(^*\)~\cite{zheng2024planagent} & 72.51 & - & 76.82 & - & - & - & - & - \\
                                                & CALMM-Drive-L\(^*\)~\cite{yao2024calmm} & 77.39 & \makecell{\colorbox{lightyellow}{88.97}} & \makecell{\colorbox{newlightgreen}{78.13}} & \makecell{\colorbox{newlightgreen}{89.71}} & 86.70 & 95.02 & 87.11 & 95.79 \\
                                                & CALMM-Drive-M\(^*\)~\cite{yao2024calmm} & 70.04 & 84.19 & 67.85 & 80.88 & 81.82 & 92.72 & 83.37 & 95.02 \\
                                       & Our Approach & 76.26 & \makecell{\colorbox{newlightgreen}{89.34}} & 76.25 & \makecell{\colorbox{newlightgreen}{89.71}} & 86.41 & \makecell{\colorbox{newlightgreen}{96.93}} & 86.37 & \makecell{\colorbox{lightyellow}{96.17}}\\
        \bottomrule
    \end{tabular}
    \label{tab:planner compare}
    \end{minipage}%
    }
    \vspace{-0.5cm}
\end{table*}

For each candidate decision $a \in \mathcal{C}$, we define a trajectory optimization objective identical to that in \cite{yao2024calmm}:
\begin{equation}
J_a = (J^f_a)^{\omega_f} \cdot (J_g)^{\omega_g},
\end{equation}
where:
\begin{itemize}
    \item $J^f_a$ is the decision-following objective, penalizing deviations from the target lane and prescribed speed interval associated with action $a$;
    \item $J_g$ is the general trajectory quality objective, instantiated as the Predictive Driver Model (PDM) scorer \cite{dauner2023parting}, which evaluates safety, comfort, and other planning-quality metrics;
    \item $\omega_f, \omega_g > 0$ balance these components during generation.
\end{itemize}
Next, decision-guided trajectory generation is performed via gradient-free diffusion-based optimization:
\begin{subequations}
\begin{align}
\mathcal{X}_a &= \text{Diffusion-ES}(J_a), \\
\mathbf{X}_a^* &= \arg\max_{\mathbf{X} \in \mathcal{X}_a} J_a(\mathbf{X}),
\end{align}
\end{subequations}
where $\text{Diffusion-ES}$ \cite{yang2024diffusion} produces a set of diverse trajectory proposals $\mathcal{X}_a$ for decision $a$, and $\mathbf{X}_a^*$ is the highest-scoring candidate. Finally, the optimal trajectory is selected via confidence-aware scoring:
\begin{subequations}\label{eq: final score}
\begin{align}
    a^\star &= \arg\max_{a \in \mathcal{C}} \Big[ \big(\hat{p}(a)\big)^{\omega_c} \cdot \tilde{J}_a(\mathbf{X}_a^*) \Big],\\
    \mathbf{X}^*_{\text{final}} &= \mathbf{X}_{a^\star}^*,
\end{align}
\end{subequations}
where $\tilde{J}_a = (J^f_a)^{\tilde{\omega}_f} \cdot (J_g)^{\tilde{\omega}_g}$ uses adjusted weights $\tilde{\omega}_f < \omega_f$ to place relatively more emphasis on general quality during final arbitration, and $\omega_c > 0$ modulates the influence of high-level uncertainty.

\subsection{Implementation Details}
\textbf{Multi-Agent Workflow.} We adopt DeepSeek-V3 to implement the action voting agent and confidence assessment agent, with \(\tau=0.7\) and \(N=10\) in the action voting stage. DeepSeek-R1 serves as summarization agent for better long-context comprehension. The workflow produces 10.2K memory items from the nuPlan training split~\cite{caesar2021nuplan} at 2\,Hz.

\textbf{Lightweight Language Model.} We employ Qwen3-1.7B as the base language model. The classifier head is a 2-layer MLP with hidden size 1024. We utilize 4-bit quantization with LoRA (rank $r = 16$) for parameter-efficient adaptation, updating only 0.08\,\% of parameters. We leverage gradient checkpointing and 8-bit AdamW for efficient training. The model is trained on four RTX-4090 (24GB) for 20 epochs ($\sim$35 hours), with \(\lambda=0.7\), \(K_{min}=0\), \(K_{max}=3\). At inference, we use 3 few-shot examples as the default setting.

\textbf{Decision-Guided Planner.} For constructing the candidate action set, we set the confidence threshold \(\gamma_c\) to 0.1. In the scoring functions, the power parameters \(\omega_f\) and \(\omega_g\) are respectively set to 5.0 and 1.0, while \(\omega_c\),  \(\tilde\omega_f\) and \(\tilde\omega_g\) are adjusted to 1.0, 0.1 and 0.3. The time resolution of motion planning is 0.1\,s, and the planning horizon is 40 time-steps. Other planner parameters are the same as in~\cite{yao2024calmm}.

\section{Experiment}
We evaluate our approach in nuPlan Test14-Random and Test14-Hard~\cite{cheng2024rethinking}, which represent normal and long-tail driving scenarios respectively. Closed-loop simulations are conducted in non-reactive (NR) and reactive (R) modes. Adopted metrics include closed-loop score (CLS) for measuring aggregated planning performance~\cite{caesar2021nuplan}, and success rate (SR) for representing robustness against severe failures (\ie, collision, wrong direction, incomplete goal, and out of drivable space)~\cite{yao2024calmm}.

\begin{figure*}
  \centering
    \includegraphics[
    width=0.87\linewidth]{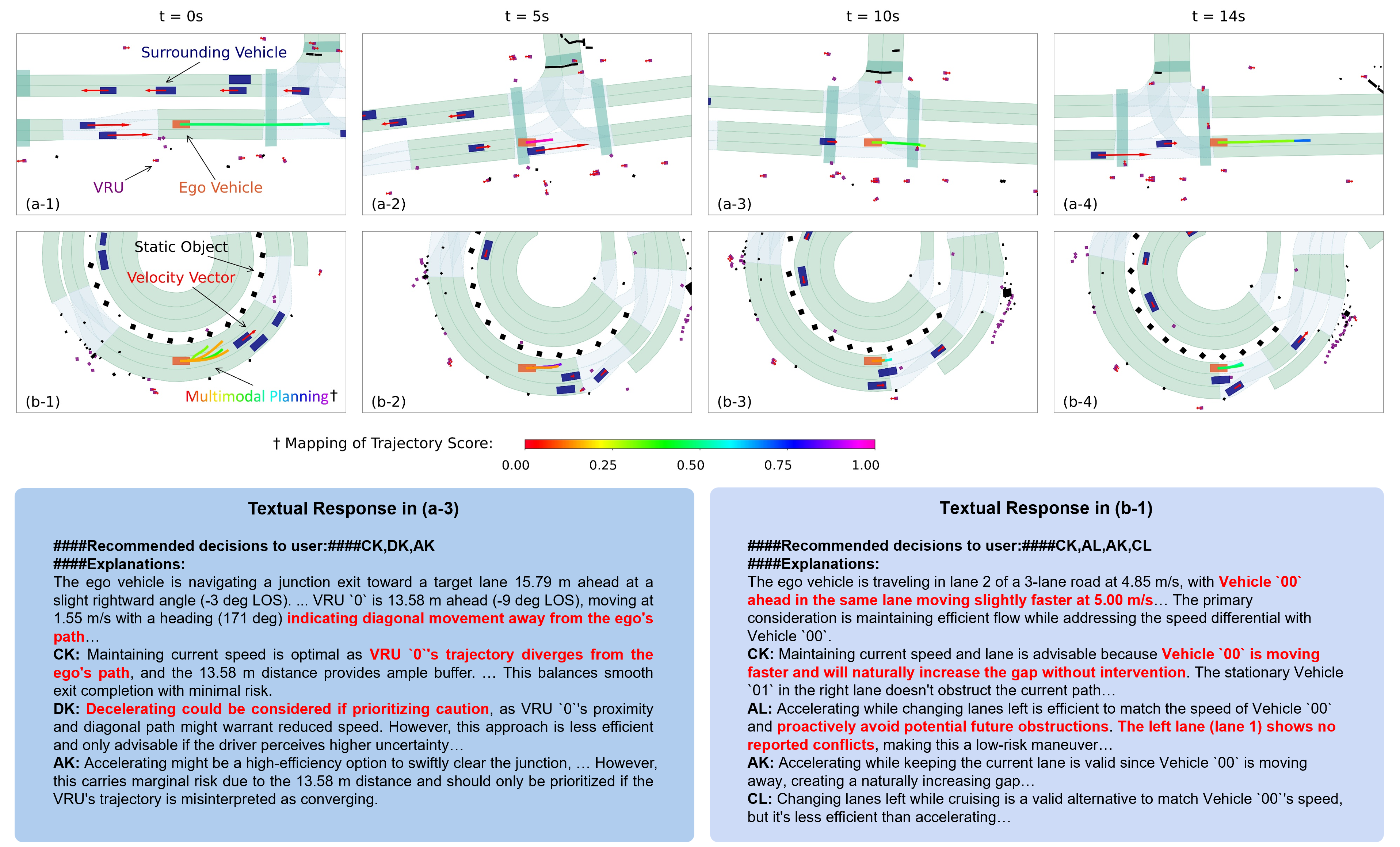}
  \caption{Qualitative demonstrations of our approach in two representative scenarios. For better visualization, the scores of different planning trajectories are normalized to sum to one after computed using Eq.~\eqref{eq: final score}. In textual responses, we highlight the key contents in interaction analysis in \textcolor{red}{red}.}
  \label{fig:qualitative}
  \vspace{-0.5cm}
\end{figure*}

\subsection{Comparative Studies}
We comparatively evaluate our approach with four categories of methods: rule-based, data-driven, hybrid (rule+data), and knowledge-driven methods. All open-source frameworks are benchmarked on a four-card RTX-4090 computational node, while the metrics of closed-source methods are quoted from original publications. 

Quantitative analysis of~\Cref{tab:planner compare} reveals four key findings: \textbf{First, our approach achieves SOTA in Test14-Hard (NR-SR 89.34\,\(\bf{\%}\), R-SR 89.71\,\(\bf{\%}\)) and Test14-Random (NR-SR 96.93\,\(\bf{\%}\)), demonstrating robust closed-loop capability.} While not topping every metric, it remains highly competitive as top results across other metrics are scattered among multiple methods of different categories. \textbf{Secondly, compared to the other lightweight knowledge-driven approach, CALMM-Drive-M, our approach shows significant improvements in almost every metric.} This validates the effectiveness of our dual-system architecture and confidence-aware fine-tuning strategy, which allows our customized lightweight model to capture reliable multimodal decision-making patterns beyond the prior knowledge of the general-purpose lightweight model.  \textbf{Thirdly, our approach balances reasoning for long-tail cases with computational tractability (\ie, single consumer-grade GPU).} While PLUTO also demonstrates competitive performance, it struggles with the reactive test in Test14-Hard due to inflexible post-processing. CALMM-Drive-L better tackles long-tail cases thanks to its common-sense reasoning ability, yet the utilization of MLLM restricts on-board deployment. \textbf{Finally, our approach surpasses data-driven planners in CLS but receives slightly lower scores than Diffusion-ES.} The latter benefits from the unmodulated PDM scorer, which mimics the nuPlan evaluator~\cite{dauner2023parting}. However, by modulating this scorer with real-time decision-making~\cite{yao2024calmm}, we achieve superior success rates (improving up to 4.41\,\(\%\)) with realistic long-term strategies while maintaining favorable planning quality.

\subsection{Qualitative Studies}\label{subsec: quali}

We showcase our framework generating multimodal driving strategies with textual explanations in two representative scenarios, as shown in Fig.~\ref{fig:qualitative}. 

\textbf{(a-1) - (a-4).} The ego vehicle is initialized on a multilane roadway and going to cross a junction. In clear traffic ahead (\(t=0\,\text{s}\)), the top-scoring trajectory performs an acceleration maneuver to allow efficient crossing of the junction. As the vehicle encounters a VRU at \(t=10\,\text{s}\), the decision-making analyzes the intention of the VRU and prioritizes a cruise maneuver since the VRU is moving away, thus balancing efficiency with safety. 

\textbf{(b-1) - (b-4).} A multimodal planning process encompassing lane-changing behaviors is visualized in the `b-1' snapshot. For high-level decision-making, `CK' and `AL' are considered the most favorable maneuvers, with `CK' being a better option since the leading vehicle is still traveling at a higher speed. After that, the leading vehicle gradually decelerates to yield to a VRU and a cut-in vehicle ahead. Then the ego vehicle changes to the left lane, allowing it to avoid obstruction and ensure travel efficiency.

\subsection{Ablation Studies}
We further investigate the performance of multiple ablated versions of our approach, which include:

\begin{itemize}
    \item \(\mathcal{M}_0\): The baseline that learns from the teacher ($\tau=0$) to infer unimodal decisions, excluding multi-agent action voting (A.V.) to explore action diversity.
    \item \(\mathcal{M}_1\): Incorporates A.V. but lacks confidence assessment (C.A.). Probabilities derived from normalized frequencies are converted to one-hot labels, utilizing an arbitrary top-category response for language modeling.
    \item \(\mathcal{M}_2\): Computes probabilities via Eq.~\eqref{eq:conf_agg} following C.A. but converts them to one-hot labels, disabling uncertainty awareness (U.A.). It uses the highest-confidence response for language modeling.
    \item \(\mathcal{M}_3\): Uses Eq.~\eqref{eq:conf_agg} probabilities for multimodal supervision. It retains the text selection logic of \(\mathcal{M}_2\), omitting the summarization (Sum.) agent for decision explanations.
    \item \(\mathcal{M}_4\): Our complete architecture integrating all proposed components.
\end{itemize}

The results in~\Cref{tab:ablation} show that our complete model achieves the best performance on all metrics. In contrast, \(\mathcal{M}_0\) with the vanilla architecture performs the worst. By incorporating action voting,  \(\mathcal{M}_1\) enables better exploration of the teacher model's reasoning space to allow the discovery of robust answers, resulting in improved closed-loop performance. Furthermore, owing to the explicit confidence assessment, \(\mathcal{M}_2\) enables a finer selection of high-quality response samples and outperforms \(\mathcal{M}_1\) on most metrics. The soft labels adopted in \(\mathcal{M}_3\) allow the lightweight model to capture the uncertainty of different decisions to facilitate efficient distillation, such that the trained model gains an evaluation closest to that of \(\mathcal{M}_4\). The remaining performance gap can be attributed to the summarization, which generates textual explanations featuring multimodal decision-making to enrich supervision signals that align with decision probabilities.

\begin{table*}[ht]
    \centering
    \resizebox{0.85\textwidth}{!}{%
    \begin{minipage}{\textwidth}
    \centering
    \caption{Results of ablation studies. All metrics are \underline{higher the better}, with the best performance highlighted in \colorbox{newlightgreen}{green}.}
    \vspace{-0.1cm}
    \footnotesize %
    \begin{tabular}{@{}p{1.0cm}cccc|cccccccc@{}}
        \toprule
        \multirow{2}{*}{\textbf{Model}} & 
        \multicolumn{4}{c|}{\textbf{Configuration}} & 
        \multicolumn{4}{c}{\textbf{Test14-Hard}} & 
        \multicolumn{4}{c}{\textbf{Test14-Random}} \\ 
        \cmidrule(lr){2-5} \cmidrule(lr){6-9} \cmidrule(lr){10-13}
        
        & \textbf{A.V.} & \textbf{C.A.} & \textbf{U.A.} & \textbf{Sum.} & 
        \textbf{NR-CLS} & \textbf{NR-SR(\%)} & \textbf{R-CLS} & \textbf{R-SR(\%)} & 
        \textbf{NR-CLS} & \textbf{NR-SR(\%)} & \textbf{R-CLS} & \textbf{R-SR(\%)} \\
        \midrule
        
        $\mathcal{M}_0$  & $\times$ & $\times$ & $\times$ & $\times$ & 73.91 & 85.66 & 72.53 & 85.66 & 84.56 & 93.87 & 83.81 & 94.64 \\
        $\mathcal{M}_1$  & \checkmark & $\times$ & $\times$ & $\times$ & 74.38 & 87.13 & 73.39 & 86.76 & 85.53 & 95.79 & 85.13 & 95.40 \\
        $\mathcal{M}_2$  & \checkmark & \checkmark & $\times$ & $\times$ & 74.37 & 86.76 & 74.98 & 88.24 & 86.01 & 96.55 & 85.33 & 95.40 \\
        $\mathcal{M}_3$  & \checkmark & \checkmark & \checkmark & $\times$ & 74.90 & 87.50 & 75.50 & 88.60 & 85.56 & 95.40 & 86.31 & \makecell{\colorbox{newlightgreen}{96.17}} \\
        $\mathcal{M}_4$ & \checkmark & \checkmark & \checkmark & \checkmark & \makecell{\colorbox{newlightgreen}{76.26}} & \makecell{\colorbox{newlightgreen}{89.34}} & \makecell{\colorbox{newlightgreen}{76.25}} & \makecell{\colorbox{newlightgreen}{89.71}} & \makecell{\colorbox{newlightgreen}{86.41}} & \makecell{\colorbox{newlightgreen}{96.93}} & \makecell{\colorbox{newlightgreen}{86.37}} & \makecell{\colorbox{newlightgreen}{96.17}} \\ %
        \bottomrule
    \end{tabular}%
    \label{tab:ablation}
    \end{minipage}%
    }
    \vspace{-0.4cm}
\end{table*}

\subsection{Effects of the Number of Few-Shots}
\begin{figure}
  \centering
    \includegraphics[
    width=0.7\linewidth]{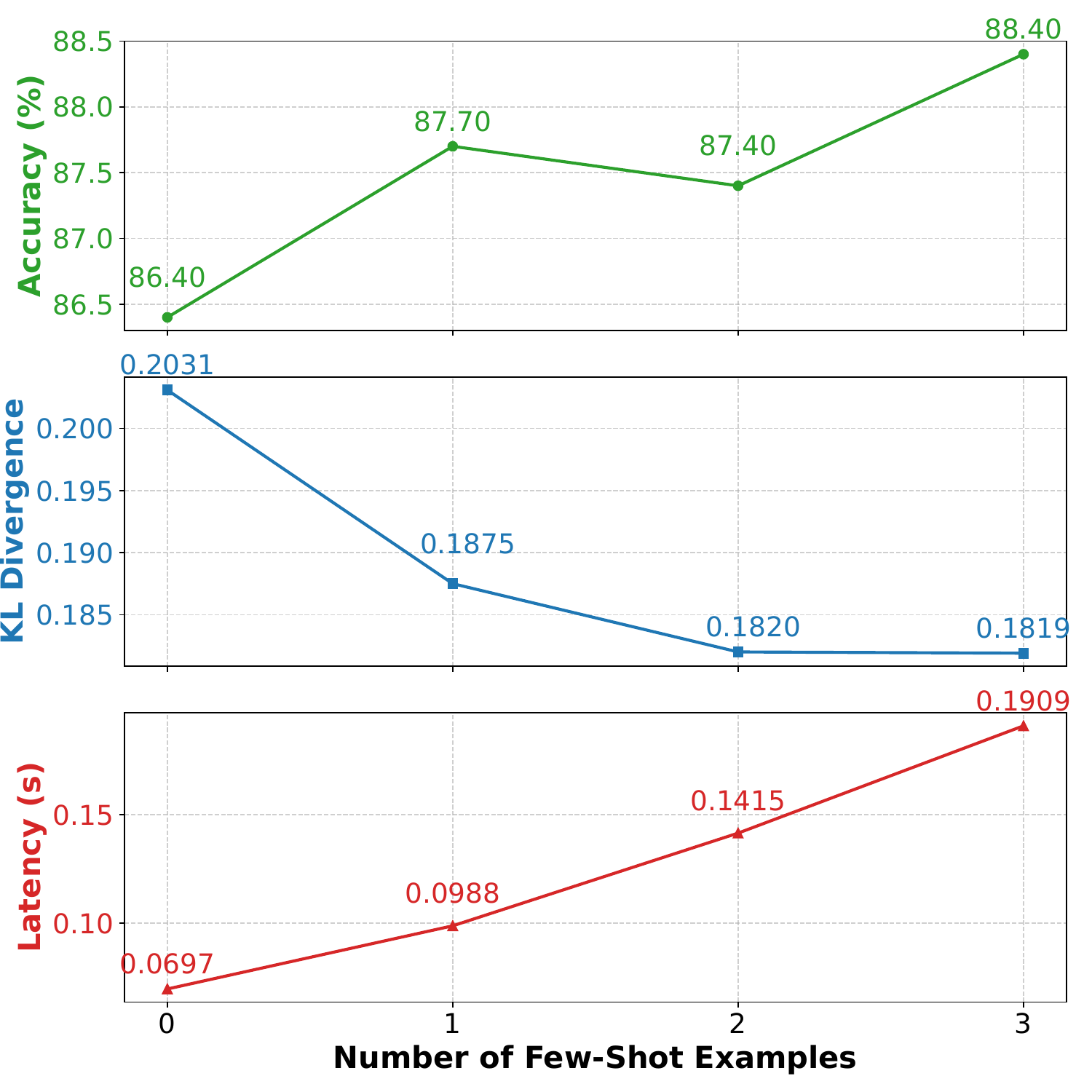}
     \vspace{-0.1cm}
  \caption{Quantitative effects of the number of few-shot examples on accuracy, KL divergence, and inference latency.}
  \label{fig:fewshot}
  \vspace{-0.6cm}
\end{figure}

We evaluate the effects of the number of few-shots in test-time RAG using 2K test samples. The metrics include:
\begin{itemize}
    \item \textit{Accuracy} (\(\uparrow\)): The percentage of times the decision with the highest probability being correctly predicted.
    \item \textit{KL Divergence} (\(\downarrow\)): The average of distance between the predicted decision probability and the true decision probability, calculated with Eq.~\eqref{eq:kld}.
    \item \textit{Latency} (\(\downarrow\)): The average inference time required to complete the decision probability prediction.
\end{itemize}
As shown in Fig.~\ref{fig:fewshot}, as the number of few-shot examples increases, the accuracy rises to 88.40\,\(\%\), while the KL divergence declines to 0.1819. This indicates that sufficient relevant contexts generally contribute to a better understanding of teachers' demonstrations. Although the latency increases from around 0.07\,s to 0.19\,s, it achieves a 26\,× speedup over competitive knowledge-driven methods (5.0\,s\,+)~\cite{zheng2024planagent} while preserving multimodal intention reasoning capabilities.

\section{Conclusion}
In this paper, we present a comprehensive decision-making framework for AD that harmonizes the reasoning depth of large foundation models with the inference efficiency required for low-latency applications. By establishing a multi-agent collaborative workflow, we synthesize a decision-making memory bank that encapsulates diverse driving intentions and uncertainty quantifications. This knowledge is effectively distilled into a lightweight language model through a novel dual-head fine-tuning objective and RAG mechanisms. Extensive closed-loop evaluations on the nuPlan platform validated the superiority of our approach. Our method not only achieves state-of-the-art success rates in challenging long-tail scenarios, but also demonstrates competitive performance against computationally expensive large model-based baselines. Future work will focus on extending this framework to end-to-end AD paradigms and validating its robustness in more diverse, real-world urban environments.

\bibliographystyle{IEEEtran}

\bibliography{main}

\end{document}